\documentclass[11pt,a4paper]{article}
\usepackage[hyperref]{acl2020}
\usepackage{times}
\usepackage{latexsym}
\usepackage{times}
\usepackage{epsfig}
\usepackage{graphicx}
\usepackage{amsmath}
\usepackage{amssymb}
\usepackage{latexsym}
\usepackage{array, multirow}
\usepackage{amsfonts}
\usepackage{mathtools}
\usepackage{rotating} 
\usepackage{rotfloat}

\usepackage{microtype}

\aclfinalcopy 

\title{Video-Grounded Dialogues with Pretrained Generation Language Models}

\author{Hung Le$^{\dag}$\thanks{\hspace{0.1cm} This work was mostly done when Hung Le was an intern at Salesforce Research Asia, Singapore.}, Steven C.H. Hoi$^{\dag\ddag}$ \\
  $^{\dag}$Singapore Management University \\
  \texttt{hungle.2018@phdcs.smu.edu.sg} \\
  $^{\ddag}$Salesforce Research Asia \\
  \texttt{shoi@salesforce.com} }

\date{}

\begin{document}
\maketitle
\begin{abstract}
Pre-trained language models have shown remarkable success in improving various downstream NLP tasks due to their ability to capture dependencies in textual data and generate natural responses. In this paper, we leverage the power of pre-trained language models for improving video-grounded dialogue, which is very challenging and involves complex features of different dynamics: (1) Video features which can extend across both spatial and temporal dimensions; and (2) Dialogue features which involve semantic dependencies over multiple dialogue turns. We propose a framework by extending GPT-2 models to tackle these challenges by formulating video-grounded dialogue tasks as a sequence-to-sequence task, combining both visual and textual representation into a structured sequence, and fine-tuning a large pre-trained GPT-2 network. Our framework allows fine-tuning language models to capture dependencies across multiple modalities over different levels of information: spatio-temporal level in video and token-sentence level in dialogue context. We achieve promising improvement on the Audio-Visual Scene-Aware Dialogues (AVSD) benchmark from DSTC7, which supports a potential direction in this line of research. 

\end{abstract}

\section{Introduction}
Recent work in large-scale pre-training transformer-based neural networks \cite{liu2019roberta, devlin-etal-2019-bert, radford2019language} has boosted the performance in various NLP tasks. The transformer-based architecture of these models allows them to capture various dependencies when trained on very large datasets. The pre-trained models are adapted into downstream tasks to generate text that is more natural, fluent, and richer than models not initialized with pre-trained weights. 
Similar to pre-trained CNN-based  neural networks developed in computer vision research \cite{he2016deep, huang2017densely} which can learn high-resolution features in images, pre-trained language models (LMs) are capable of capturing fine-grain textual dependencies in text data of rich semantics. 

While the benefits of pre-trained language models present in many downstream NLP tasks such as machine translation and question answering (QA) \cite{devlin-etal-2019-bert, lan2019albert}, 
they are particularly suitable to adapt to dialogue response generation tasks for two major reasons: 
(1) Dialogue response generation usually involves more complex dynamics between input and output text sequences. The input typically involves dialogue history, including conversational exchanges between users and dialogue agents. A dialogue agent needs to capture relevant dependencies along each dialogue turns to generate a sensible response. 
(2) Compared to other NLP tasks, it is very challenging to collect and create large-scale dialogue datasets. Adopting pre-training approaches could ameliorate the limited dialogue datasets by leveraging rich linguistic dependencies learned from other available text data. 
We are motivated by these observations to adapt pre-trained language models into a dialogue task and improve the quality of generated responses.

\begin{figure*}[htb]
	\centering
	\resizebox{1.0\textwidth}{!} {
	\includegraphics{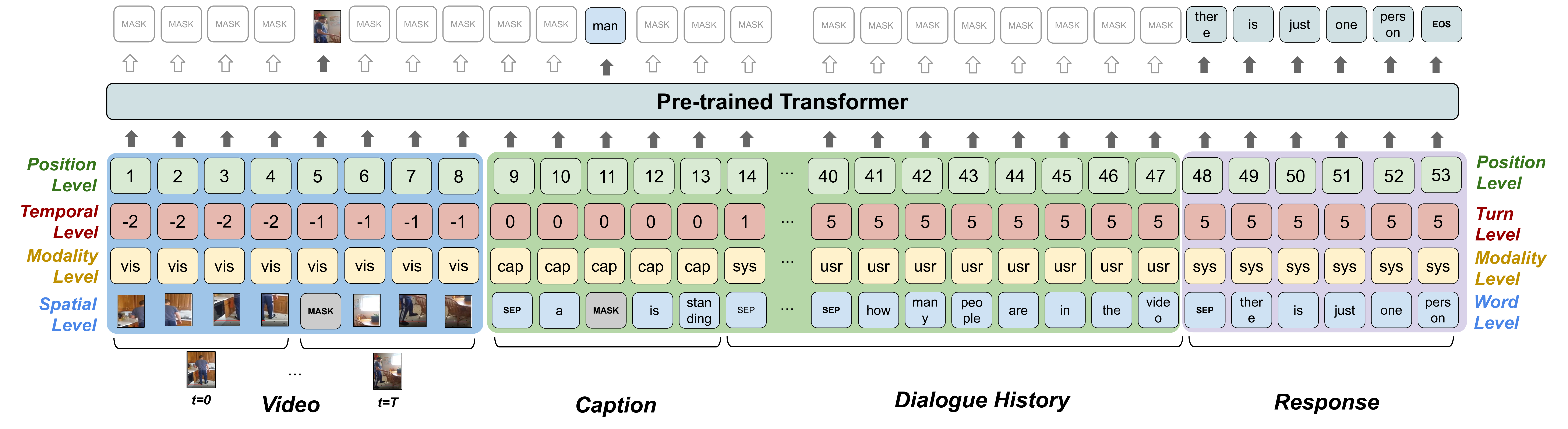}
	}
	\caption{The proposed VGD-GPT2 architecture for video-grounded dialogues based on the pre-trained transformer model (GPT-2). The video and text input are combined together over multiple encoding layers to inject different attributes to encoded features.}
	\label{fig:model}
     \vspace{-0.1in}
\end{figure*}

Along the line of research that combines both vision and language \cite{antol2015vqa, hori2019avsd}, transformer-based neural networks can also be applied to capture various dependencies across different types of input modalities (text and image) with appropriate objective loss functions \cite{alberti-etal-2019-fusion, su2019vl, chen2019uniter}. 
The multi-head attention mechanism of these models can detect long-range dependencies between each token in the input text and each image patch or spatial objects in the input image. 
We extend this framework to a video-dialogue task and fully leverage the power of pre-trained models to obtain linguistic and visual representations in dialogues and videos. 
Specifically, we tackle the Audio-Visual Scene-aware Dialogues (AVSD) task \cite{hori2019avsd} which aims to generate dialogue responses grounded on both visual and audio features of the video.
The dialogue agent needs to create responses that not only match the dialogue flow but also address user questions about a given video over multiple dialogue turns. 

First, we detail how to formulate input components of a video-grounded dialogue as a downstream task of pre-trained language models. 
We follow the general sequence-to-sequence framework, whereby the input components are combined to a structured sequence of multiple modalities and output is a system response.
We then apply pre-trained models \cite{radford2019language} to leverage the deep attention neural networks to capture text and video dependencies with fine granularity.
Specifically, we propose to capture dependencies between each token in text data and each spatial feature along the temporal dimension of the input video. Lastly, we present a multi-task learning framework that includes additional learning objectives in addition to dialogue response generation objective.
Our promising results on the AVSD benchmark demonstrate the efficacy of our proposed framework.

\section{Related Work}
We briefly describe related work in two major lines of research: dialogues and vision-text modeling. 

\subsection{Dialogue Modeling}
\citet{whang2019domain} applies pre-trained language models for response selection tasks in open-domain dialogues.
The output of the language model (e.g. $\mathrm{[CLS]}$ token in BERT) is used as a contextual representation of each pair of dialogue context and candidate response.
\citet{budzianowski-vulic-2019-hello} assumes access to ground-truth dialogue states and generates responses in task-oriented dialogues by combining input components into a single sequence. As dialogue states and database states are used as raw text input, the models can be fine-tuned from a deep pre-trained language model such as GPT.
\citet{chao2019bert} and \citet{lai2020simple} use pre-trained LMs to track dialogue states in task-oriented dialogues by utilizing the output representations to predict slot values. 
In this work, we aim to address video-grounded dialogue tasks and generate natural responses in an end-to-end manner. 

\subsection{Vision-Text Modeling}
The transformer-based neural architecture of pre-trained language models has been used to learn cross-modal representations for vision-text NLP tasks. 
\citet{li2019unicoder} uses a BERT-based architecture to improve linguistic and visual representations for image captioning tasks. 
\citet{lu2019vilbert} follows a similar approach to tackle visual QA but segregates the visual and text input components rather combining both into a single sequence.
\citet{alberti-etal-2019-fusion} leverages a pre-trained BERT model to improve cross-modal representations in either early fusion or late fusion approach. 
We are motivated to extend this line of research to a video-based setting. Video is considered much more complicated than image due to the additional temporal variation across video frames. 
A related work to ours is VideoBERT \cite{Sun_2019_ICCV} which utilizes BERT models for video captioning. Instead of using visual features to represent video frames, VideoBERT transforms frame-level features into visual tokens and uses them as raw text input to a BERT-based architecture. 

\section{Method}
Our model architecture can be seen in Figure \ref{fig:model}.
We are inspired by Transformer-based LM approaches that leverage different levels of features in text, such as word, character, and position levels. 
We apply this principle and technique to overcome the challenge in AVSD which involves multi-turn dialogue input combined with video input with spatial-temporal variations. 
We propose to decompose videos into patches but maintain a structured temporal sequence. This sequence is then directly combined with text inputs of dialogue which are also arranged in a temporally ordered manner. This kind of feature reformulation is simple yet powerful as it allows explicit dependency learning across all pairs of text tokens and video patches. 
Therefore, it can facilitate stronger signals to answer human queries in greater granularities.

\subsection{Model Architecture}
We trained a GPT model based on the GPT-2 \cite{radford2019language} architecture. The GPT-2 model is based on the transformer network \cite{vaswani17attention} which includes 12 to 24 layers of masked multi-head attention on very large text data. 
Following the success of GPT-2 in generation-based tasks, we adapt the power of GPT-2 pre-trained models to generate video-grounded dialogue responses and call our framework ``VGD-GPT2".
First, we modify the input components as a long sequence of video frames or video segments and dialogue turns.\\

\noindent \textbf{Video Representations}. Each video frame or video segment is further structured as a sequence of spatial regions, which can be extracted using a pre-trained video model. 
For an input video $V$, we denote the output of a pre-trained 2D CNN or 3D CNN video model as $Z^\mathrm{pre}_\mathrm{V} \in \mathbb{R}^{F \times P \times d_\mathrm{emb}}$ where $d_\mathrm{emb}$ is the feature dimension of the pre-trained video model, $F$ is the resulting number of sampled video frames or video segments, and $P$ is the number of spatial regions in each video frame. 
We reshape $Z_\mathrm{V}$ as a sequence of image patches and pass it through a linear transformation with ReLU activation to match the feature dimension $d$ of pre-trained language model:
\begin{align}
    Z^\mathrm{spatial}_V = \mathrm{ReLU} (Z^\mathrm{pre}_{V} W_V) \in \mathbb{R}^{F P \times d}
\end{align}
where $W_V \in \mathbb{R}^{d_\mathrm{emb} \times d}$. We denote this as \emph{spatial-level} features of input video.
As can be seen from Figure \ref{fig:model}, we inject different types of input attributes into $X_V$ by adding three additional encoding layers:\\
(1) \emph{Modality-level} encoding that informs the type of information. We use a modality token ``vis" to uniformly represent visual information type.\\
(2) \emph{Temporal-level} encoding that informs model the frame-level (or segment-level) position of input features.\\
(3) \emph{Position-level} encoding that incorporates the spatial-level ordering. This is equivalent to the positional encoding of tokens in sentences seen in BERT-based language models.\\
All the three layers are trainable parameters to enable models to learn the dynamics of input features. All encoding layers are modeled to have the same feature dimension $d$ of the pre-trained model. 
We combine all encoding layers through element-wise summation, resulting in a rich video representation:
\begin{align}
Z_V = Z^\mathrm{spatial}_\mathrm{V}+Z^\mathrm{mod}_\mathrm{V}+Z^\mathrm{temporal}_\mathrm{V}+Z^\mathrm{pos}_\mathrm{V}
\end{align}
\noindent \textbf{Text Representations}. Similarly, we break down dialogue history $H$ as sequence of dialogue turns $H = (H_1, H_2, ..., H_t)$ where $t$ is the current dialogue turn. Each dialogue turn is represented as a pair of user utterance $U$ and system response $S$ concatenated sequentially $H=((U_1, S_1), (U_2, S_2), ..., U_{t}))$ ($S_{t}$ is the target response that need to be generated by the models). 
Each utterance is then represented as a sequence of tokens $x$ so the dialogue history can be represented as $X_H=(x_1, x_2, ..., x_{L_H})$ and $Y=S_t=(y_1, y_2, ..., y_{L_Y})$ where $L_H$ and $L_Y$ are the total number of tokens in the dialogue history and target response respectively. 
Following the AVSD setting \cite{hori2019avsd}, we utilize the text input of video caption $C$. The video caption typically provides a linguistic summary of the video in one or two sentences. The caption can be represented as a sequence of tokens $X_C=(x_1, x_2, ..., x_{L_C})$.
We combine all text input sequences to form a single sequence $X_\mathrm{T}=(X_C,X_H,Y_{-1})$ as input to the models.
$Y_{-1}$ is the target response sequence shifted left by one position to enable auto-regressive prediction of output tokens. 
We denote embedded features as $Z^\mathrm{token}_\mathrm{T}$ as the \emph{token-level} encoding layer of the text input.
Similar to video features, we add additional layers to inject different attributes of $X_\mathrm{T}$ (See Figure \ref{fig:model}):\\
(1) \emph{Modality-level} encoding that differentiates segments in $X_\mathrm{T}$. We use 3 different modality tokens: ``cap", ``sys", and ``usr" to specify whether the token in the corresponding position is part of input caption, system responses, or user utterances.\\
(2) \emph{Turn-level} encoding that encodes the turn number of the token in the corresponding position.\\
(3) \emph{Position-level} encoding that is used to inject signals of the token ordering.\\

Similar to video representation, the encoded input is combined through element-wise summation:
\begin{align}
Z_T = Z^\mathrm{token}_\mathrm{T}+Z^\mathrm{mod}_\mathrm{T}+Z^\mathrm{turn}_\mathrm{T}+Z^\mathrm{pos}_\mathrm{T}
\end{align}
We concatenated both $Z_V$ and $Z_T$ to create a single input sequence $Z_{VT}$ of length $(F \times P + L_C + L_H + L_Y)$ and embedding dimension $d$. $Z_{VT}$ is used as input to a pre-trained GPT-2 for fine-tuning. 

\subsection{Optimization}
\label{subsec:optim}
Following a similar strategy adopted by \citet{wolf2019transfertransfo}, we fine-tune the models in a multi-task setting with the following objectives:

\noindent(1) \textit{Response Generation}: this is a typical objective function that maximizes the likelihood of output target response conditioned on the source sequence. 

\noindent(2) \textit{Masked Multi-modal Modeling}: we explore two loss functions: masked language modeling (MLM) and masked visual modeling (MVM). We mask both tokens and spatial regions in video frames in training instances and require the model to re-generate them with the remaining inputs.
MLM is learned similarly as response generation by passing through a linear layer with softmax. MVM is learned by minimizing the L1 loss in feature space between the output representation of the masked visual region and the original input representation. 
Both are passed through a linear transformation to the same dimensional space. 
This is similar to the perceptual loss proposed by \cite{johnson2016perceptual, dosovitskiy2016generating} for image style transfer and image resolution tasks.
We follow BERT \cite{devlin-etal-2019-bert} and replace about 15\% of tokens and image region inputs in each training instance at random with a \emph{[MASK]} token. The corresponding output representations are then used to recover the original tokens or image regions.

\noindent(3) \textit{Matching Video-Text Pair} (MVT): for about 15\% of training instances, we adapt the pretrained language model to the dialogue domain by replacing the original input with an incorrect dialogue or video input at random. We use a special token \emph{[CLS]} concatenated to the input sequence to learn the contextual representation. The vector integrates contextual cues through Transformer attention layers and the corresponding output representation is used to predict if the input video-text pair is correct. 

\section{Experiments}
\subsection{Experimental Testbed and Setup}
We use the open-source implementation of the GPT-2 architecture and obtain pre-trained model checkpoints \footnote{\url{https://github.com/huggingface/transfer-learning-conv-ai}}. 
We experiment with two pre-trained GPT-2 models: small (S) and medium (M) \cite{radford2019language}.
We use Adam optimizer with a learning rate of 5e-5 based on grid search. 
We adopt a learning rate decay schedule as similarly used by \citet{vaswani17attention}. 
we set the weight on the response generation loss to be 1.5 times higher than the other losses. 

We experiment with the the video-grounded dialogue task in the large-scale AVSD benchmark in DSTC7 \cite{hori2019avsd}. The AVSD benchmark contains dialogues grounded on the Charades videos \cite{sigurdsson2016hollywood}. Each dialogue consists of up to 10 dialogue turns, each turn including a user utterance and system response (See Table \ref{tab:datasets} for more details of the dataset). 

To extract visual features, we used the 3D CNN-based ResNext-101 \cite{xie2017aggregated} pre-trained on Kinetics \cite{hara2018can} to obtain the spatio-temporal video features. We fixed the batch size to 16 and the maximum sequence length compatible with the corresponding GPT2 models. We sampled video features every 16 frames without overlapping. We trained up to 50 epochs on 4 GPUs. 
We report the objective scores, including BLEU, METEOR, ROUGE-L, and CIDEr. 
We compare system-generated responses with 6 reference ground-truth responses. 

\begin{table}[htbp]
	\centering
	\begin{tabular}{llll}
    \hline
    \textbf{\#}     & \multicolumn{1}{c}{\textbf{Train}} & \multicolumn{1}{c}{\textbf{Val.}} & \multicolumn{1}{c}{\textbf{Test}} \\ \hline
    Dialogs & 7,659                              & 1,787                             & 1,710                             \\ \ 
     Turns   & 153,180                            & 35,740                            & 13,490                            \\ 
     Words   & 1,450,754                          & 339,006                           & 110,252                           \\ \hline
    \end{tabular}
	\caption{Summary of DSTC7 AVSD.}
	\label{tab:datasets}
	\vspace{-0.2in}
\end{table}
\begin{table*}[htbp]
\resizebox{1.0\textwidth}{!} {
\begin{tabular}{lccccclllllll}
\hline
\multicolumn{1}{c}{\textbf{Model}} & \textbf{Spatial} & \textbf{Temporal} & \textbf{MLM} & \textbf{MVM} & \textbf{MVT} & \multicolumn{1}{c}{\textbf{BLEU1}} & \multicolumn{1}{c}{\textbf{BLEU2}} & \multicolumn{1}{c}{\textbf{BLEU3}} & \multicolumn{1}{c}{\textbf{BLEU4}} & \multicolumn{1}{c}{\textbf{METEOR}} & \multicolumn{1}{c}{\textbf{ROUGE-L}} & \multicolumn{1}{c}{\textbf{CIDEr}} \\
\hline
Baseline                       &                  & \checkmark                 &              &              &              & 0.626                              & 0.485                              & 0.383                              & 0.309                              & 0.215                               & 0.487                                & 0.746                              \\
AVSD Winner                       &                  & \checkmark                 &              &              &              & 0.718                              & 0.584                              & 0.478                              & 0.394                              & 0.267                               & 0.563                                & 1.094                              \\
MTN                                &                  & \checkmark                 &              &              &              & 0.731                              & 0.597                              & 0.490                              & 0.406                              & 0.271                               & 0.564                                & 1.127                              \\  
VGD-GPT2 (S)                           & \checkmark                & \checkmark                 & \checkmark            & \checkmark            & \checkmark            & 0.750                              & \textbf{0.621}                     & {0.516}                     & 0.433                     & \textbf{0.283}                      & {0.581}                       & \textbf{1.196}                     \\
VGD-GPT2 (S)                           & \checkmark                &                   & \checkmark            & \checkmark            & \checkmark            & \textbf{0.753}                     & 0.619                              & 0.512                              & 0.424                              & 0.280                               & 0.571                                & 1.185                              \\
VGD-GPT2 (S)                           &                  & \checkmark                 & \checkmark            & \checkmark            & \checkmark            & 0.750                              & 0.616                              & 0.511                              & 0.427                              & 0.280                               & 0.579                                & 1.188                              \\
VGD-GPT2 (S)                           & \checkmark                & \checkmark                 & \checkmark            & \checkmark            &              & 0.745                              & 0.613                              & 0.508                              & 0.423                              & 0.281                               & 0.579                                & 1.173                              \\
VGD-GPT2 (S)                           & \checkmark                & \checkmark                 & \checkmark            &              & \checkmark            & 0.749                              & 0.613                              & 0.505                              & 0.419                              & 0.274                               & 0.571                                & 1.153                              \\
VGD-GPT2 (S)                           & \checkmark                & \checkmark                 &              & \checkmark            & \checkmark            & 0.744                              & 0.612                              & 0.505                              & 0.421                              & 0.281                               & {0.581}                       & 1.192                              \\
VGD-GPT2 (M)                           & \checkmark                & \checkmark                 & \checkmark            & \checkmark            & \checkmark            &   0.749                                 &  0.620                                  &  \textbf{0.520}                                  & \textbf{0.436}                                   &    0.282                                 &     \textbf{0.582}                                 &   {1.194}                     \\
\hline
\end{tabular}
}
\caption{Evaluation on the AVSD benchmark of baselines and different variants of VGD-GPT2 based on: (1) video features in spatial or temporal (or both) dimension and (2) fine-tuning objective functions: MLM - masked language modeling, MVM: mask visual modeling, and MVT - matching video-text pair.}
\label{tab:results}
\vspace{-0.2in}
\end{table*}

\subsection{Results}
We compare the proposed VGD-GPT2 model with the following baseline models:\\
(1) \textbf{Baseline} \cite{hori2019avsd} proposes a novel sequence-to-sequence approach with question-guided LSTM on both video visual and audio temporal features. Dialogue history is encoded by a hierarchical LSTM and the final representation is concatenated with question and video representations as input to decode dialog responses.\\
(2) \textbf{AVSD Winner} \cite{sanabria2019cmu} extends the previous work with more refined visual features and transfer learning from a video summary task.\\
(3) \textbf{MTN} \cite{le-etal-2019-multimodal} adopts a transformer-based approach with question-guided attention on visual features formulated as an auto-encoding module. Table \ref{tab:results} shows the details of our results. 

Our VGD-GPT2 model outperforms the existing approaches across all the automated metrics. The results show that fine-tuning a language model with video-grounded dialogues can help to generate quality responses and improve model performance. 
By initializing our models with a language model pre-trained on massive text data, we obtain richer feature representations that capture more complex dependencies between inputs. 

Compared with the baseline with Transformer-based neural networks \cite{le-etal-2019-multimodal}, our model treats both visual and text features with equal importance at different levels of different dimensions. 
Specifically, we aligned the token level with spatial level and turn level with temporal level between visual and text features. 
By contrast, MTN only considers the temporal variation of the visual features and mainly focuses on text-based attention. 
Our early fusion strategy with a multi-level alignment approach of multi-modal inputs allows higher resolution relations between all feature representations in later layers of neural networks.
\vspace{-0.1in}
\subsection{Ablation Analysis}
Besides, Table \ref{tab:results} also shows that fine-tuning a pre-trained model with both spatial-temporal information and multi-task objectives can benefit the main task of response generation. 
To obtain spatial-only and temporal-only features, we follow a similar approach from \cite{jang2017tgif} by using average pooling to pool the visual features along the temporal or spatial dimensions. 
Considering CIDEr as the evaluation measure, learning dependencies in both spatial and temporal dimensions can improve the performance by 0.01 absolute score from spatial-only feature and 0.008 absolute score from temporal-only feature. 

Our proposed auxiliary objectives also help to improve model performance by adapting the pre-trained model to the current data domain, video-based dialogues. 
MLM and MVM are used to improve learning of local dependencies in token and spatial levels, while MVT is used to support learning global dependencies between text and visual modalities. 
We observe that adding MVM objective function can increase the CIDEr score the most, by 0.043 absolute score, as compared to adding MVT (0.023 absolute score) or MLM (0.004 absolute score) objective function. 

We also found moderate performance improvements in BLEU3, BLEU4, and ROUGE-L, when increasing GPT-2 from small to medium size.
We note that the increasing model parameters in GPT-2 may require longer fine-tuning procedure or a larger dialogue training dataset to fully optimize the models in the dialogue domain.
\vspace{-0.1in}
\section{Conclusions}
\vspace{-0.1in}
In this work, we leverage pre-trained language models for a video-grounded dialogue task. We propose a sequence-to-sequence framework and a multi-task fine-tuning approach to adapt the pre-trained models to the video dialogue domain. Despite using GPT-2 models, our framework can be extended with other language models and similarly adopted to improve other multi-modal dialogues.
Our early fusion strategy effectively unifies different levels of features in both dialogues and video without complicating the network architecture.


\bibliography{acl2020}

\begin{thebibliography}{28}
\expandafter\ifx\csname natexlab\endcsname\relax\def\natexlab#1{#1}\fi

\bibitem[{Alberti et~al.(2019)Alberti, Ling, Collins, and
  Reitter}]{alberti-etal-2019-fusion}
Chris Alberti, Jeffrey Ling, Michael Collins, and David Reitter. 2019.
\newblock \href {https://doi.org/10.18653/v1/D19-1219} {Fusion of detected
  objects in text for visual question answering}.
\newblock In \emph{Proceedings of the 2019 Conference on Empirical Methods in
  Natural Language Processing and the 9th International Joint Conference on
  Natural Language Processing (EMNLP-IJCNLP)}, pages 2131--2140, Hong Kong,
  China. Association for Computational Linguistics.

\bibitem[{Antol et~al.(2015)Antol, Agrawal, Lu, Mitchell, Batra,
  Lawrence~Zitnick, and Parikh}]{antol2015vqa}
Stanislaw Antol, Aishwarya Agrawal, Jiasen Lu, Margaret Mitchell, Dhruv Batra,
  C~Lawrence~Zitnick, and Devi Parikh. 2015.
\newblock Vqa: Visual question answering.
\newblock In \emph{Proceedings of the IEEE international conference on computer
  vision}, pages 2425--2433.

\bibitem[{Budzianowski and Vuli{\'c}(2019)}]{budzianowski-vulic-2019-hello}
Pawe{\l} Budzianowski and Ivan Vuli{\'c}. 2019.
\newblock \href {https://doi.org/10.18653/v1/D19-5602} {Hello, it{'}s {GPT}-2 -
  how can {I} help you? towards the use of pretrained language models for
  task-oriented dialogue systems}.
\newblock In \emph{Proceedings of the 3rd Workshop on Neural Generation and
  Translation}, pages 15--22, Hong Kong. Association for Computational
  Linguistics.

\bibitem[{Chao and Lane(2019)}]{chao2019bert}
Guan-Lin Chao and Ian Lane. 2019.
\newblock Bert-dst: Scalable end-to-end dialogue state tracking with
  bidirectional encoder representations from transformer.
\newblock \emph{arXiv preprint arXiv:1907.03040}.

\bibitem[{Chen et~al.(2019)Chen, Li, Yu, Kholy, Ahmed, Gan, Cheng, and
  Liu}]{chen2019uniter}
Yen-Chun Chen, Linjie Li, Licheng Yu, Ahmed~El Kholy, Faisal Ahmed, Zhe Gan,
  Yu~Cheng, and Jingjing Liu. 2019.
\newblock Uniter: Learning universal image-text representations.
\newblock \emph{arXiv preprint arXiv:1909.11740}.

\bibitem[{Devlin et~al.(2019)Devlin, Chang, Lee, and
  Toutanova}]{devlin-etal-2019-bert}
Jacob Devlin, Ming-Wei Chang, Kenton Lee, and Kristina Toutanova. 2019.
\newblock \href {https://doi.org/10.18653/v1/N19-1423} {{BERT}: Pre-training of
  deep bidirectional transformers for language understanding}.
\newblock In \emph{Proceedings of the 2019 Conference of the North {A}merican
  Chapter of the Association for Computational Linguistics: Human Language
  Technologies, Volume 1 (Long and Short Papers)}, pages 4171--4186,
  Minneapolis, Minnesota. Association for Computational Linguistics.

\bibitem[{Dosovitskiy and Brox(2016)}]{dosovitskiy2016generating}
Alexey Dosovitskiy and Thomas Brox. 2016.
\newblock Generating images with perceptual similarity metrics based on deep
  networks.
\newblock In \emph{Advances in neural information processing systems}, pages
  658--666.

\bibitem[{Hara et~al.(2018)Hara, Kataoka, and Satoh}]{hara2018can}
Kensho Hara, Hirokatsu Kataoka, and Yutaka Satoh. 2018.
\newblock Can spatiotemporal 3d cnns retrace the history of 2d cnns and
  imagenet?
\newblock In \emph{Proceedings of the IEEE conference on Computer Vision and
  Pattern Recognition}, pages 6546--6555.

\bibitem[{He et~al.(2016)He, Zhang, Ren, and Sun}]{he2016deep}
Kaiming He, Xiangyu Zhang, Shaoqing Ren, and Jian Sun. 2016.
\newblock Deep residual learning for image recognition.
\newblock In \emph{Proceedings of the IEEE conference on computer vision and
  pattern recognition}, pages 770--778.

\bibitem[{{Hori} et~al.(2019){Hori}, {Alamri}, {Wang}, {Wichern}, {Hori},
  {Cherian}, {Marks}, {Cartillier}, {Lopes}, {Das}, {Essa}, {Batra}, and
  {Parikh}}]{hori2019avsd}
C.~{Hori}, H.~{Alamri}, J.~{Wang}, G.~{Wichern}, T.~{Hori}, A.~{Cherian}, T.~K.
  {Marks}, V.~{Cartillier}, R.~G. {Lopes}, A.~{Das}, I.~{Essa}, D.~{Batra}, and
  D.~{Parikh}. 2019.
\newblock \href {https://doi.org/10.1109/ICASSP.2019.8682583} {End-to-end audio
  visual scene-aware dialog using multimodal attention-based video features}.
\newblock In \emph{ICASSP 2019 - 2019 IEEE International Conference on
  Acoustics, Speech and Signal Processing (ICASSP)}, pages 2352--2356.

\bibitem[{Huang et~al.(2017)Huang, Liu, Van Der~Maaten, and
  Weinberger}]{huang2017densely}
Gao Huang, Zhuang Liu, Laurens Van Der~Maaten, and Kilian~Q Weinberger. 2017.
\newblock Densely connected convolutional networks.
\newblock In \emph{Proceedings of the IEEE conference on computer vision and
  pattern recognition}, pages 4700--4708.

\bibitem[{Jang et~al.(2017)Jang, Song, Yu, Kim, and Kim}]{jang2017tgif}
Yunseok Jang, Yale Song, Youngjae Yu, Youngjin Kim, and Gunhee Kim. 2017.
\newblock Tgif-qa: Toward spatio-temporal reasoning in visual question
  answering.
\newblock In \emph{Proceedings of the IEEE Conference on Computer Vision and
  Pattern Recognition}, pages 2758--2766.

\bibitem[{Johnson et~al.(2016)Johnson, Alahi, and
  Fei-Fei}]{johnson2016perceptual}
Justin Johnson, Alexandre Alahi, and Li~Fei-Fei. 2016.
\newblock Perceptual losses for real-time style transfer and super-resolution.
\newblock In \emph{European conference on computer vision}, pages 694--711.
  Springer.

\bibitem[{Lai et~al.(2020)Lai, Tran, Bui, and Kihara}]{lai2020simple}
Tuan~Manh Lai, Quan~Hung Tran, Trung Bui, and Daisuke Kihara. 2020.
\newblock A simple but effective bert model for dialog state tracking on
  resource-limited systems.
\newblock In \emph{ICASSP 2020-2020 IEEE International Conference on Acoustics,
  Speech and Signal Processing (ICASSP)}, pages 8034--8038. IEEE.

\bibitem[{Lan et~al.(2020)Lan, Chen, Goodman, Gimpel, Sharma, and
  Soricut}]{lan2019albert}
Zhenzhong Lan, Mingda Chen, Sebastian Goodman, Kevin Gimpel, Piyush Sharma, and
  Radu Soricut. 2020.
\newblock \href {https://openreview.net/forum?id=H1eA7AEtvS} {Albert: A lite
  bert for self-supervised learning of language representations}.
\newblock In \emph{International Conference on Learning Representations}.

\bibitem[{Le et~al.(2019)Le, Sahoo, Chen, and Hoi}]{le-etal-2019-multimodal}
Hung Le, Doyen Sahoo, Nancy Chen, and Steven Hoi. 2019.
\newblock \href {https://doi.org/10.18653/v1/P19-1564} {Multimodal transformer
  networks for end-to-end video-grounded dialogue systems}.
\newblock In \emph{Proceedings of the 57th Annual Meeting of the Association
  for Computational Linguistics}, pages 5612--5623, Florence, Italy.
  Association for Computational Linguistics.

\bibitem[{Li et~al.(2019)Li, Duan, Fang, Jiang, and Zhou}]{li2019unicoder}
Gen Li, Nan Duan, Yuejian Fang, Daxin Jiang, and Ming Zhou. 2019.
\newblock Unicoder-vl: A universal encoder for vision and language by
  cross-modal pre-training.
\newblock \emph{arXiv preprint arXiv:1908.06066}.

\bibitem[{Liu et~al.(2019)Liu, Ott, Goyal, Du, Joshi, Chen, Levy, Lewis,
  Zettlemoyer, and Stoyanov}]{liu2019roberta}
Yinhan Liu, Myle Ott, Naman Goyal, Jingfei Du, Mandar Joshi, Danqi Chen, Omer
  Levy, Mike Lewis, Luke Zettlemoyer, and Veselin Stoyanov. 2019.
\newblock Roberta: A robustly optimized bert pretraining approach.
\newblock \emph{arXiv preprint arXiv:1907.11692}.

\bibitem[{Lu et~al.(2019)Lu, Batra, Parikh, and Lee}]{lu2019vilbert}
Jiasen Lu, Dhruv Batra, Devi Parikh, and Stefan Lee. 2019.
\newblock Vilbert: Pretraining task-agnostic visiolinguistic representations
  for vision-and-language tasks.
\newblock In \emph{Advances in Neural Information Processing Systems}, pages
  13--23.

\bibitem[{Radford et~al.(2019)Radford, Wu, Child, Luan, Amodei, and
  Sutskever}]{radford2019language}
Alec Radford, Jeff Wu, Rewon Child, David Luan, Dario Amodei, and Ilya
  Sutskever. 2019.
\newblock Language models are unsupervised multitask learners.

\bibitem[{Sanabria et~al.(2019)Sanabria, Palaskar, and Metze}]{sanabria2019cmu}
Ramon Sanabria, Shruti Palaskar, and Florian Metze. 2019.
\newblock Cmu sinbad’s submission for the dstc7 avsd challenge.
\newblock In \emph{DSTC7 at AAAI2019 workshop}, volume~6.

\bibitem[{Sigurdsson et~al.(2016)Sigurdsson, Varol, Wang, Farhadi, Laptev, and
  Gupta}]{sigurdsson2016hollywood}
Gunnar~A Sigurdsson, G{\"u}l Varol, Xiaolong Wang, Ali Farhadi, Ivan Laptev,
  and Abhinav Gupta. 2016.
\newblock Hollywood in homes: Crowdsourcing data collection for activity
  understanding.
\newblock In \emph{European Conference on Computer Vision}, pages 510--526.
  Springer.

\bibitem[{Su et~al.(2020)Su, Zhu, Cao, Li, Lu, Wei, and Dai}]{su2019vl}
Weijie Su, Xizhou Zhu, Yue Cao, Bin Li, Lewei Lu, Furu Wei, and Jifeng Dai.
  2020.
\newblock \href {https://openreview.net/forum?id=SygXPaEYvH} {Vl-bert:
  Pre-training of generic visual-linguistic representations}.
\newblock In \emph{International Conference on Learning Representations}.

\bibitem[{Sun et~al.(2019)Sun, Myers, Vondrick, Murphy, and
  Schmid}]{Sun_2019_ICCV}
Chen Sun, Austin Myers, Carl Vondrick, Kevin Murphy, and Cordelia Schmid. 2019.
\newblock Videobert: A joint model for video and language representation
  learning.
\newblock In \emph{The IEEE International Conference on Computer Vision
  (ICCV)}.

\bibitem[{Vaswani et~al.(2017)Vaswani, Shazeer, Parmar, Uszkoreit, Jones,
  Gomez, Kaiser, and Polosukhin}]{vaswani17attention}
Ashish Vaswani, Noam Shazeer, Niki Parmar, Jakob Uszkoreit, Llion Jones,
  Aidan~N Gomez, \L~ukasz Kaiser, and Illia Polosukhin. 2017.
\newblock \href
  {http://papers.nips.cc/paper/7181-attention-is-all-you-need.pdf} {Attention
  is all you need}.
\newblock In I.~Guyon, U.~V. Luxburg, S.~Bengio, H.~Wallach, R.~Fergus,
  S.~Vishwanathan, and R.~Garnett, editors, \emph{Advances in Neural
  Information Processing Systems 30}, pages 5998--6008. Curran Associates, Inc.

\bibitem[{Whang et~al.(2019)Whang, Lee, Lee, Yang, Oh, and
  Lim}]{whang2019domain}
Taesun Whang, Dongyub Lee, Chanhee Lee, Kisu Yang, Dongsuk Oh, and HeuiSeok
  Lim. 2019.
\newblock Domain adaptive training bert for response selection.
\newblock \emph{arXiv preprint arXiv:1908.04812}.

\bibitem[{Wolf et~al.(2019)Wolf, Sanh, Chaumond, and
  Delangue}]{wolf2019transfertransfo}
Thomas Wolf, Victor Sanh, Julien Chaumond, and Clement Delangue. 2019.
\newblock Transfertransfo: A transfer learning approach for neural network
  based conversational agents.
\newblock \emph{arXiv preprint arXiv:1901.08149}.

\bibitem[{Xie et~al.(2017)Xie, Girshick, Doll{\'a}r, Tu, and
  He}]{xie2017aggregated}
Saining Xie, Ross Girshick, Piotr Doll{\'a}r, Zhuowen Tu, and Kaiming He. 2017.
\newblock Aggregated residual transformations for deep neural networks.
\newblock In \emph{Proceedings of the IEEE conference on computer vision and
  pattern recognition}, pages 1492--1500.

\end{thebibliography}
\bibliographystyle{acl_natbib}




\end{document}